\documentclass[10pt,twocolumn,letterpaper]{article}

\usepackage{cvpr}              
\usepackage[table,dvipsnames]{xcolor}
\usepackage{colortbl}
\definecolor{mygrey}{rgb}{0.9,0.9,0.9}
\usepackage{graphicx}
\usepackage{amsmath}
\usepackage{amssymb}
\usepackage{booktabs}
\usepackage{bm}
\usepackage{mathtools}
\usepackage{algpseudocode}
\usepackage{algorithm}
\usepackage{siunitx}
\usepackage{multirow}
\usepackage{enumitem}
\usepackage{caption}
\usepackage{diagbox}
\usepackage[dvipsnames]{xcolor}
\definecolor{myGreen}{RGB}{0,114,0}
\usepackage[pagebackref,breaklinks,colorlinks,citecolor=myGreen]{hyperref}

\usepackage[capitalize]{cleveref}
\crefname{section}{Sec.}{Secs.}
\Crefname{section}{Section}{Sections}
\Crefname{table}{Table}{Tables}
\crefname{table}{Tab.}{Tabs.}

\def\paperID{5207} 
\def\confName{CVPR}
\def\confYear{2024}

\graphicspath{ {images/} }

\algnewcommand\algorithmiclocalize{\textbf{Localize neuron and update model:}}
\algnewcommand\Localize{\item[\algorithmiclocalize]}
\algnewcommand\algorithmiccross{\textbf{Cross-section plane determination:}}
\algnewcommand\Cross{\item[\algorithmiccross]}
\algnewcommand\algorithmicbranchc{\textbrowcolorf{Bifurcation candidates detection:}}
\algnewcommand\Branchc{\item[\algorithmicbranchc]}

\newcommand{\diag}[1]{\text{diag}}
\newcommand{\conj}[1]{\text{conj}}

\makeatletter

\newcommand{\bimtdp}[0]{\texttt{Bi-MTDP}}
\newcommand{\bimtdpA}[0]{\texttt{Bi-MTDP-C}}
\newcommand{\bimtdpB}[0]{\texttt{Bi-MTDP-F}}

\begin{document}

\title{Efficient Multitask Dense Predictor via Binarization}

\author{
  \textbf{Yuzhang Shang$^{1}$, Dan Xu$^2$, Gaowen Liu$^{3}$, Ramana Rao Kompella$^{3}$,}
  \textbf{Yan Yan$^{1,}$\footnotemark[2]}\\
  $^{1}$Illinois Institute of Technology,  $^{2}$HKUST, $^{3}$Cisco Research\\
\tt\small{yzshawn@outlook.com, danxu@cse.ust.hk, \{gaoliu, rkompell\}@cisco.com, yyan34@iit.edu}
}

\maketitle

\renewcommand{\thefootnote}{\fnsymbol{footnote}}
\footnotetext[2]{Corresponding author}

\begin{abstract}
Multi-task learning for dense prediction has emerged as a pivotal area in computer vision, enabling simultaneous processing of diverse yet interrelated pixel-wise prediction tasks. 
However, the substantial computational demands of state-of-the-art (SoTA) models often limit their widespread deployment. 
This paper addresses this challenge by introducing network binarization to compress resource-intensive multi-task dense predictors. 
Specifically, our goal is to significantly accelerate multi-task dense prediction models via Binary Neural Networks (BNNs) while maintaining and even improving model performance at the same time. 
To reach this goal, we propose a Binary Multi-task Dense Predictor, \bimtdp, and several variants of \bimtdp, in which a multi-task dense predictor is constructed via specified binarized modules. 
Our systematical analysis of this predictor reveals that performance drop from binarization is primarily caused by severe information degradation. To address this issue, we introduce a deep information bottleneck layer that enforces representations for downstream tasks satisfying Gaussian distribution in forward propagation. Moreover, we introduce a knowledge distillation mechanism to correct the direction of information flow in backward propagation. 
Intriguingly, one variant of \bimtdp~outperforms full-precision (FP) multi-task dense prediction SoTAs, ARTC~\cite{bruggemann2021exploring} (CNN-based) and InvPT~\cite{ye2022inverted} (ViT-Based). This result indicates that \bimtdp~is not merely a naive trade-off between performance and efficiency, but is rather a benefit of the redundant information flow thanks to the multi-task architecture. Code is available at \href{https://github.com/42Shawn/BiMTDP}{BiMTDP}.
\end{abstract}
\section{Introduction}
\label{sec:intro}
\begin{figure}
    \centering
    \includegraphics[width=0.48\textwidth]{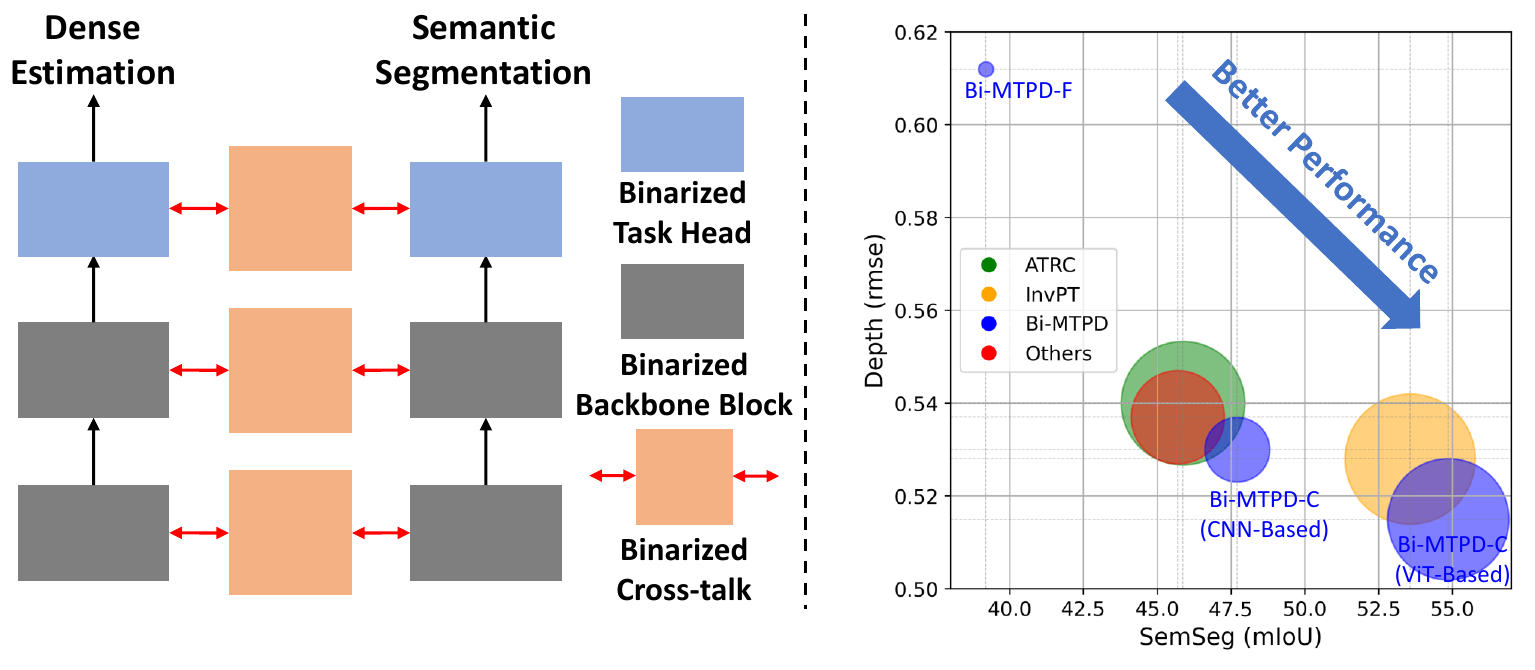}
    \vspace{-0.2in}
    \captionsetup{font=small}
    \caption{(\textbf{Left}) A conceptual illustration of binary dense predictions in a multi-task manner. In contrast to approaching a series of relevant tasks individually, the multitask model benefits from information supplementation among different tasks via cross-talk structures, but the cumbersome cross-talk modules also add additional computational burden. (\textbf{Right}) Performance summary on NYUD-v2. X-axis and Y-axis denote the performance on depth estimation (lower is better) and segmentation (higher is better), respectively. Size of dots denotes FLOPs. ATRC~\cite{bruggemann2021exploring} and InvPT~\cite{ye2022inverted} are previous CNN-based and ViT-based SoTAs, respectively.}
    \vspace{-0.2in}
    \label{fig:intro_brief}
\end{figure}

There is a growing trend in the computer vision community where dense prediction tasks are processed in a multi-task learning manner, such as semantic segmentation, monocular depth estimation, and human parsing~\cite{misra2016cross,xu2018pad,vandenhende2020mti,vandenhende2021mtlsurvey}. Benefited from the information supplementation mechanism via cross-talk structures in the multi-task models, the overall performance for the series of dense prediction tasks has been greatly improved~\cite{kendall2018multi} (see Fig.~\ref{fig:intro_brief}). 
However, the computational demands of State-of-the-Art (SoTA) multi-task dense prediction models, which process multiple complex pixel-wise tasks concurrently, are substantial. This high computational requirement limits their application in resource-constrained environments like autonomous driving, robotics, and virtual reality. Our goal is to optimize these heavy SoTA models for edge devices, balancing speed with performance.

Several strategies for neural network compression have been explored, including pruning~\cite{han2015deep}, network quantization~\cite{hubara2016binarized,shang2023causal,shang2023post} and knowledge distillation~\cite{hinton2015distilling}. 
Notably, network binarization, a form of quantization, minimizes weights and activations to $\pm 1$, enabling the replacement of computationally intensive inner-product operations in full-precision networks with more efficient xnor-bitcount operations in Binary Neural Networks (BNNs)~\cite{hubara2016binarized}. 
Binarization theoretically reduces memory costs by 32$\times$ and increases inference speed by 64$\times$, making BNNs suitable for edge-device.

While BNNs have shown impressive results in image classification, achieving nearly full-precision ResNet-level accuracy~\cite{liu2020reactnet}, their application has largely been limited to small-scale models, overlooking other computationally intensive computer vision tasks~\cite{hubara2016binarized,qin2020forward,liu2020reactnet,liu2020bi,shang2022network}.
Extending BNNs to larger models should be the next step.
However, this expansion has been hindered by issues such as overfitting~\cite{kim2021improving,shang2022lipschitz,shang2022network} and information degradation~\cite{qin2021bibert}.
Techniques effective in full-precision models, like label smoothing~\cite{reichel2009simple}, dropout~\cite{srivastava2014dropout}, and mixup~\cite{zhang2017mixup} have less effect on BNNs~\cite{kim2020binaryduo,shang2022lipschitz,shang2022network}.
Furthermore, SoTA multi-task dense prediction tasks often require deep and complex models, equipped with multi-modality fusion structures~\cite{vandenhende2020mti,bruggemann2021exploring,ye2022inverted}, exacerbating the challenges in implementing binarization effectively.

The primary barrier to applying binarization in multi-task dense prediction tasks is the significant degradation of information flow in deep models~\cite{hubara2016binarized,qin2020forward,qin2021bibert}, leading to reduced performance. 
To address this issue, we first propose a Binary Multitask Dense Predictor (\bimtdp) baseline, where a multi-task dense predictor is formulated via binarized modules.
Based on a thorough review of this baseline, we conclude that the binarization operation destroys the information flows in multi-task models, and thus representations for downstream tasks are not informative compared with their full-precision counterparts. To tackle this problem, we update \bimtdp~with additional information flow calibration mechanisms in two directions. 
First, we implement variational information bottleneck enforcing the embeddings to follow Gaussian distribution with sparsity in forward propagation, in order to filter out the task-irrelevant factors. Second, we leverage the existing FP models via feature-based knowledge distillation to calibrate the gradient of the binary network in backward pass. 

The benefits of \bimtdp~can be analyzed from two orthogonal perspectives. On one hand, from the perspective of network binarization, the accomplishment of bridging binarization with the multitask dense prediction framework testifies that \bimtdp~can effectively supplement information, and consequently improve the performance of the individual binary models. 
On the other hand, from the perspective of multitask dense prediction task, accelerating those cumbersome models is profitable to design more effective and efficient cross-talk modules in it, as shown in Fig.~\ref{fig:intro_brief}. Since existing dense prediction models have severe limitation in modelling the cross-talk modules due to their heavy utilization of convolution operation, it is critical for the multitask dense predictions to learn interactions and inference covering various scopes of the multitask context via the cross-talk mechanism~\cite{misra2016cross,xu2018pad,zhang2019pattern,vandenhende2020mti,bruggemann2021exploring,ye2022inverted}. 
Intriguingly, a variant of \bimtdp~ outperforms SoTA approach ATRC~\cite{bruggemann2021exploring} by 4\% over the segmentation task while remaining 43\% faster in speed, implying that our proposed method is not a naive trade-off between performance and efficiency. By empirically investigating this ``free lunch'' achievement, we conclude that the win-win outcome is benefited from our designed information supplementation mechanism which strengthens the representation ability of the binary model.

\section{Related Work}
\label{related}

\noindent\textbf{Multitask Dense Prediction.}
Multi-Task Learning (MTL) methods can be generally categorized into two main paradigms in terms of the way where model learns shared representations: hard and soft parameter sharing. Hard parameter sharing characterizes architectures which typically share the first hidden representations among the tasks while branching to independent task-specific representations at a later stage. Most approaches split to task-specific heads at a single branch point~\cite{kokkinos2017ubernet,chen2018gradnorm,kendall2018multi,sener2018multi}. However, such naive branching can be sub-optimal, raising interest in mechanisms that allow for finely branched architectures~\cite{lu2017fully,vandenhende2019branched}. 
As a result, in soft parameter sharing, each task is assigned its own set of parameters and a feature sharing mechanism realize the cross-talk as demonstrated in Fig~\ref{fig:intro_brief}. The following works devise the cross-talk mechanisms focusing on the locations in the network where information or features are exchanged or shared between tasks. Apart from the locations, the feature sharing modules are also widely studied. For example, feature fusing can be introduced along the entire network depth~\cite{misra2016cross,gao2019nddr}; PAD-Net~\cite{xu2018pad} uses multi-modal distillation to enhance task-specific predictions, in which information flow from each source to target task is regulated with a sigmoid-activated gate function; and MTI-Net ~\cite{vandenhende2020mti} combines the multi-modal distillation module of PAD-Net with a multi-scale refinement scheme to facilitate cross-task talk at multiple scales.

Although increasing the number of cross-talk modules intuitively benefits the overall performance of the models, computational cost is often an obstacle. To handle this issue, ATRC~\cite{bruggemann2021exploring} introduces NAS~\cite{zoph2016neural} to automatically design an efficient information fusing modules. From the perspective of the efficient representation cross-talk, our proposed models with the binarization module can be interpreted as a new pathway to feature fusing within a notably lower inference speed level.

\noindent\textbf{Neural Network Binarization.}
As pioneers, \cite{hubara2016binarized} use the sign function to quantize weights and activations to $\pm 1$, initiating the trends of studies of network binarization. To tackle the vanishing gradient issue induced by the binarization operations, the straight-through estimator (STE)~\cite{bengio2013estimating} is introduced for the gradient approximation. Rooted in this archetype, considerable studies contribute to improving the performance of BNNs, particularly on ImageNet. For example, \cite{liu2020bi} propose Bi-Real introducing double residual connections with FP downsampling layers to mitigate the excessive gradient vanishing issue caused by binarization, and consequently demonstrate that delicately designing additional connections within BNNs benefits the gradient propagation. \cite{he2020proxybnn} design a proxy matrix as a basis of the latent parameter space to guide the alignment of the weights with different bits by recovering the smoothness of BNNs. In summary, a large number of methods have extended the boundary of the network binarization \wrt accuracy over classification (\eg, ReActNet~\cite{liu2020reactnet} within comparable FLOPs of binary ResNet-18 achieves 65.9\% Top-1 accuracy on ImageNet, while full-precision version is only 70.5\%). 

However, most of those works validate their effectiveness over classification with relatively small architectures (mostly ResNet18 and ResNet34). Meanwhile, the network-based models for dense prediction tasks are bigger and deeper than those toy models, as the information flow in networks is severely degraded. Consequently, directly implementing existing binarization methods can not achieve supposed success. To mitigate the degradation, we propose to binarize those dense prediction models in a multitask way. 

\section{Multitask Network Binarization}
\subsection{Binary Neural Network}
To begin with, we briefly review the general idea of binary neural networks (BNNs) in~\cite{courbariaux2015binaryconnect, hubara2016binarized}. Here, we only elaborate the speedup mechanism and the degradation of information flow of the binarization. We define a full-precision (FP) neural network with $K$ layers, $f(\mathbf{x}) = (\mathbf{W}^{K}\times\sigma\times \mathbf{W}^{K-1} \cdots \sigma\times \mathbf{W}^{1})(\mathbf{x})$, where $\mathbf{x}$ is the input sample and $\mathbf{W}^{k}:\mathbb{R}^{d_{k-1}} \longmapsto \mathbb{R}^{d_{k}}(k=1,...,K)$ stands for the weight matrix connecting the $(k-1)$-th and the $k$-th layer, with $d_{k-1}$ and $d_{k}$ representing the sizes of the input and output of the $k$-th network layer, respectively. The $\sigma(\cdot)$ function performs element-wise activation for the feature maps. 

\begin{figure}[h]
\centering
  \includegraphics[width=0.46\textwidth]{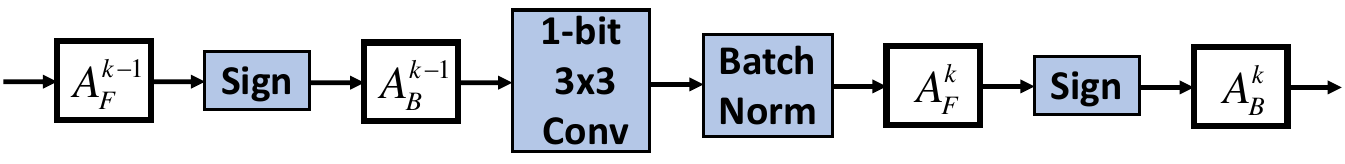}
  \captionsetup{font=small}
  \caption{A general illustration of the forward propagation of the $k$-th layer in the BNN.}
  \vspace{-0.1in}
  \label{fig:bnn}
\end{figure}

BNNs vary from FP neural networks in terms of the forward operation and the backward gradient approximation. Specifically, in the forward propagation, the BNN maintains FP latent weights $\mathbf{W}_F$ for gradient updates, and the $k$-th weight matrix $\mathbf{W}_F^k$ is binarized into $\pm 1$, obtaining the binary weight matrix $\mathbf{W}_B^k$ via the binarize function $\textit{sign}(\cdot)$, \textit{i.e.} $\mathbf{W}_B^k = \textit{sign}(\mathbf{W}_F^k)$. Then the intermediate activation map (full-precision) of the $k$-th layer is produced by $\mathbf{A}_F^{k} = \mathbf{W}_B^k \mathbf{A}_B^{k-1}$. The same quantization method is used to binarize the full-precision activation map as $\mathbf{A}_B^k = \textit{sign}(\mathbf{A}_F^k)$, and the whole forward pass of binarization is performed by iterating this process for $L$ times, as shown in Fig.~\ref{fig:bnn}. For BNNs, the weights and activations are 1-bit, by which the network is accelerated 32 times in terms of memory cost. Importantly, inference of BNN is accelerated 64 times, as the FP multiplication in FP networks is replaced by Xnor-Bitcount in BNNs.

In the backward propagation, the main challenge is that the pervasive sign functions are theoretically non-differentiable, and thus extremely destroy the information flow via gradient the propagation. To address this problem, a large number of researchers~\cite{qin2020binary} widely exploit the straight-through estimator (STE)~\cite{bengio2013estimating} to numerically approximate the derivative of the whole BNN, \textit{i.e.}
\begin{equation}
\text{Backward:~~~} \frac{\partial\mathcal{L}}{\partial x}= \left\{
    \begin{array}{ll}
        \frac{\partial\mathcal{L}}{\partial \textit{sgn}(x)} & \quad \vert x \vert \leq 1 \\
        0 & \quad \vert x \vert > 1.
    \end{array}
    \right.
    \label{eq:critic_new}
\end{equation}
It is worth noting that we do not implement the aforementioned vanilla approximation method in practice, while we utilize the prevalent Bi-Real~\cite{liu2020bi} and IR-Net~\cite{qin2020forward} to gradually approximate the \textit{sign} function, which have been proven to be better estimation approaches~\cite{Qin:iclr21,qin2021bibert}.

Even though numerous methods have been proposed to eliminate the deterioration of the information flow induced by the binarization, the deterioration is still inevitable due to the severe accuracy loss of weights, activations and gradients~\cite{Qin:iclr21,qin2021bibert}.
Consequently, binarization destroys the performance of the complicated computer vision models. 

\begin{figure*}[!t]
\centering
  \includegraphics[width=0.9\textwidth]{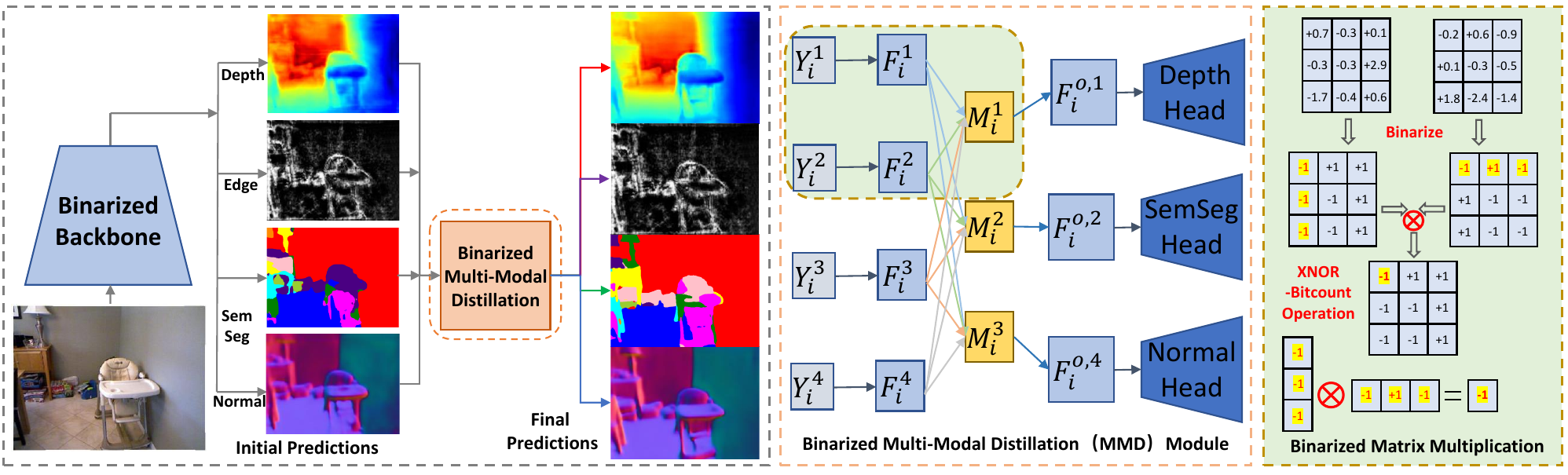}
  \captionsetup{font=small}
  \caption{\textbf{(Left)} The illustration of the baseline multitask framework. \textbf{(Middle)} The designed MMD modules for binary representations. Importantly, the MMD module can pass information among different predictions, acting as a cross-talk mechanism. \textbf{(Right)} As all fundamental modules in \bimtdp~baseline are binarized, inferences can be performed by complete Bool operations, which are very computationally cheap.}
  \vspace{-0.1in}
  \label{fig:mtl}
\end{figure*}

\subsection{Multitask Dense Predictor} 
After deploying binarization techniques in the models for dense prediction tasks, the performance of the binarized models is unacceptable, as shown in Fig.~\ref{fig:intro_brief} and the binary single result in Table.~\ref{table:ablation}. Since the architectures of those SoTA dense prediction models are relatively heavier and deeper (\eg, HRNet-48 or ResNet-101 with a task-specific head~\cite{maninis2019attentive}) than the ones for classification (\eg, ResNet-18 with a fully-connected layer as the classification head). Moreover, the information passing in the binary models via back-propagation, especially in deep models, is notoriously inefficient~\cite{liu2020bi}. 

Dense prediction tasks can mutually supplement information, \eg, surface normals and depth can directly be derived from each other, which can be modeled as the regularization of each other~\cite{vandenhende2020mti}. The relevancy among dense prediction tasks is worth being utilized to improve the overall performance of models. For example, before the deep learning era, pioneering work~\cite{gupta2014learning} utilizes RGB-D images with depth information to predict scene semantics to improve the quality of the prediction. In the deep learning era, recent attention-based multitask learning methods~\cite{xu2018pad,vandenhende2020mti,bruggemann2021exploring} explicitly and implicitly distill information from other tasks as a complementary signal to improve the targeted task performance. Briefly, the above-mentioned methods are achieved by combining an existing backbone network for initial task predictions with a subsequent decoding process, as shown in Fig.~\ref{fig:mtl} (Left).

Specifically, the shared features extracted by the backbone network are then processed by a set of task-specific heads, which produce a series of initial predictions for $T$ tasks, \textit{i.e.} $\{Y_i^k\}~(k = 1,\cdots, T)$ (the backbone and the task-specific heads are referred as the front-end of the network~\cite{vandenhende2020mti}). 
Transforming and binarizing $Y_i^t$ into the form of a 1-bit feature map, we obtain a set of corresponding binary feature maps of the scene, \ie $\{F_{B,i}^t\}~(t = 1,\cdots,T)$ which are more task-aware than the shared binary features of the backbone network. 
The information from these task-specific feature representations is then fused via a \textit{multi-modal distillation via binarized attention mechanism} before making the final task predictions. As previous work featured, our method is also task-saleable. Especially, it is possible that some tasks are only predicted in the front-end of the network (initial prediction). The initial tasks are also called auxiliary tasks since they serve as proxies in order to improve the performance of the final tasks, as shown in Fig.~\ref{fig:mtl}.

\noindent\textbf{Multi-Modal Distillation (MMD) via Binarized Attention Mechanism.} The multi-modal distillation module is the key in the multi-task dense prediction model. Specifically, we utilize the attention mechanism for guiding the information passing between the binary feature maps generated from different modalities for different tasks. Since the passed information flow is not always helpful, the attention can act as a gate function to control the flow, in other words, to make the network automatically learn to focus or to ignore information from other binary features~\cite{xu2018pad,vandenhende2020mti}. Including the binarization operations, we can formalize the MMD via binarized attention as follows. While passing information to the $k$-th task, we first obtain a binarized attention map $\mathbf{A}_{B,i}^k$, \ie,
\begin{equation}
    \mathbf{A}_{B,i}^{k} \longleftarrow \textit{bool}(\mathbf{W}_{B}^{k}\otimes\mathbf{F}_{B,i}^{k})
\end{equation}
where $\mathbf{W}_{B}^{k}$ is the parameters of the binarized convolution layer, $\mathbf{F}_{B,i}^{k}$ is the binary feature map of the initial prediction, and $\otimes$ denotes convolution operation. Then the information is passed with the attention map controlled as follows:
\begin{equation}
    \mathbf{F}_{B,i}^{o,k} \longleftarrow \textit{sign}\left[ \mathbf{F}_{B,i}^{k} + \sum_{t=1, t\neq k}^{T}\mathbf{A}_{B,i}^k\odot(\mathbf{W}_{B,t}\otimes \mathbf{F}_{B,i}^{t})\right]
\end{equation}
where $\odot$ element-wise multiplication. The general demonstration of the distillation process is presented in Fig.~\ref{fig:mtl} Left. The output binary feature map $\mathbf{F}_{B,i}^{o,k}$ is then used by the head for the corresponding $t$-th task in Fig.~\ref{fig:mtl} Right. By using the task-specific distillation activations, the network can preserve more information for each task~\cite{xu2018pad,vandenhende2020mti,bruggemann2021exploring}, which especially benefits the BNNs where the deteriorated information flow mainly induce the performance drop.

\begin{figure*}[!h]
\centering
  \includegraphics[width=0.90\textwidth]{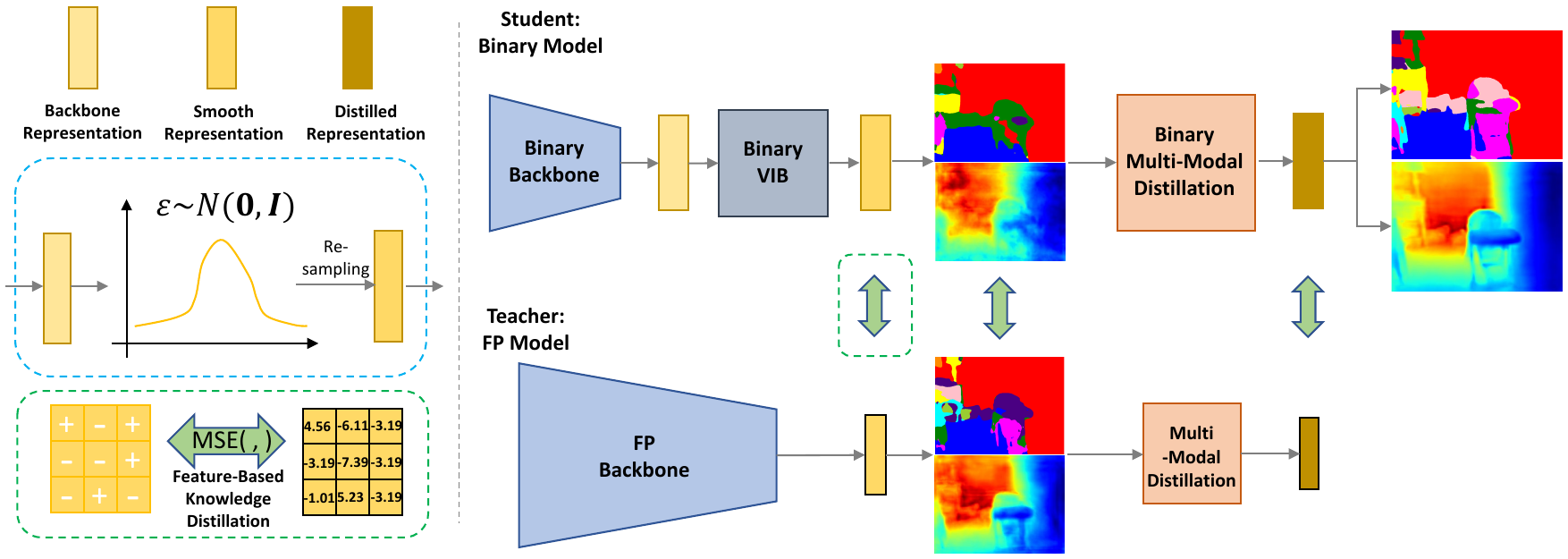}
  \captionsetup{font=small}
  \caption{\textbf{The pipeline of \bimtdp.} We introduce a VIB layer after the backbone network to filter-out the nuisance factors which may lead to model overfitting issue in the forward propagation. In addition, we deploy a feature-based knowledge distillation mechanism to guide the optimization direction in the backward propagation.}
  \vspace{-0.1in}
  \label{fig:intuition}
\end{figure*}

On the other hand, multitask dense prediction models benefit from the network binarization in terms of performance. Although these multitask models have achieved a promising performance, they are still limited by the nature of convolution-based distillation modules that are heavily used in a multi-scale way, which model critical spatial and task-related interactions in relatively local perceptive fields~\cite{ye2022inverted}. Theoretically, more distillation modules in different network nodes can contribute to model performance, yet we cannot unrestrictedly add distillation modules to the existing model due to the computational limitation. Fortunately, with the saved computational cost of the binary networks, we can implement additional distillation modules in our model. 

\noindent\textbf{Binary Baseline for Multitask Dense Prediction, \bimtdp.}
To obtain dense predictions with BNNs under a multitask learning framework, we practically binarize the MTI-Net~\cite{vandenhende2020mti} as the binary baseline. Specifically, the main modules in the full-precision MTI-Net, including backbone, heads, and multi-modal distillation module, are replaced with binary modules (both weights and activations are 1-bit). We call this baseline as \bimtdp.

\subsection{Information Flow Supplementation}

Although we build a fully binarized baseline \bimtdp~for multitask dense predictions and train the pipeline with common techniques, the performance is still of major concern. The baseline suffers an immense information degradation as the \textit{nuisance factors are over-fitted in the forward propagation} and \textit{optimization directions severely mismatch in the backward propagation}. To solve these problems, in this section, we further propose the variant of \bimtdp, \bimtdpB. Specifically, we introduce a variational information bottleneck (VIB) layer after the output of the shared binary backbone to precisely enforces the feature extractor to preserve the minimal sufficient information of the input data. As well, we deploy the feature-based knowledge distillation to guided the optimization direction. We present more details in the following section.

\noindent\textbf{Variational Information Bottleneck for Filter-Out Nuisance Factors.}
Obtaining the initial binary representations of input images by the shared backbone, we need to train a series of targeted heads to split them. A straightforward strategy is to feed these representations into the following MMD module. However, the binarized representations lack homogenization leading to model overfitting issue~\cite{wang2020bidet}. Therefore, the need to regularize the binarized representations, while the regularization would not to contaminate the information flow in the representations. Fortunately, the information bottleneck (IB) principle directly relates to compression with the best hypothesis that the data misfit and the model complexity should simultaneously be minimized~\cite{tishby2000information,wang2020bidet}. 

As the VIB could effectively capture the relevant parts and filter out the irrelevant ones from inputs, we design a novel VIB-based layer after the backbone. In particular, it explicitly enforces the feature extractor to preserve the minimal sufficient information of the input data. In other words, it can help ensure the information flow flexibly to learn clean representation for the targeted tasks. The objective function of our VIB-based classification can be formulated as a term of information loss, written as follows: 
\begin{equation}
    \mathcal{L}_{vib} = \textit{KL}\left[p(\mathbf{Z}\mid \mathbf{A_{B}}), r(\mathbf{Z}) \right],
\end{equation}
where $\mathbf{A_{B}}$ is the input binary backbone representation,  $\mathbf{B}$ is the latent representation variable, $p(\mathbf{Z}\mid \mathbf{A_{B}})$ is a multivariate Gaussian distribution, and $r(\mathbf{Z})$ is a standard normal distribution. Generally, the latter is a regularization term controlling how much information of the input is filtered out. A more detailed discussion about the VIB for binarized models filtering out irrelevant information is in Supplemental Materials.

\begin{figure*}[t]
\centering
  \includegraphics[width=0.83\textwidth]{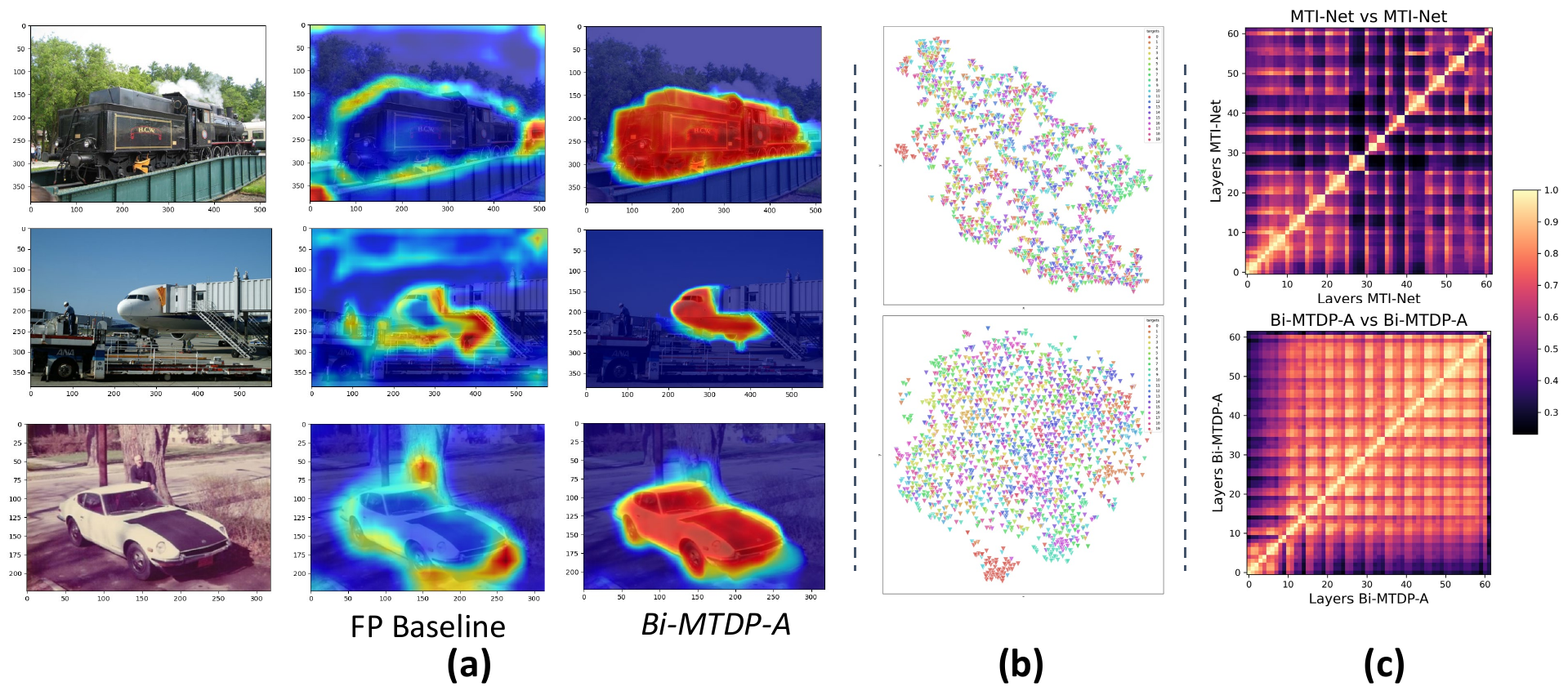}
  \captionsetup{font=small}
  \caption{\textbf{(a)} Grad-cam visualization of feature maps of different multitask dense prediction methods. \textbf{(b)} t-SNE visualization of learned features of all 20 classes on Pascal-Context. \textbf{(c)} Centered Kernel Alignment analyzing the information flow within different networks.}
  \vspace{-0.2in}
  \label{fig:representation}
\end{figure*}

\noindent\textbf{Feature-based Knowledge Distillation for Guiding the Direction of Information Flow.}
Distillation is a common and essential optimization approach to alleviate the performance drop of quantized models on ultra-low bit-width settings, which can be flexibly deployed for any architectures to utilize the knowledge of a full-precision teacher model~\cite{qin2021bibert,jiao2019tinybert,bulat2020high,shang2021london}. The usual practice is to distill the activations in a layerwise manner from the full-precision teacher to the quantized counterparts, \textit{i.e.}, $\mathbf{F}_{B,l}$ and $\mathbf{F}_{FP,l}$ ($l = 1, \cdots, L$, where $L$ represents the number of network layers), respectively. We use the mean squared errors (MSE) as the distance function to measure the difference between corresponding from features student and teacher. The knowledge distillation loss can be written as follows:
\begin{equation}
    \mathcal{L}_{kd} = \sum_{l=1}^{L}\textit{MSE}(\mathbf{F}_{B,l},\mathbf{F}_{FP,l}).
\end{equation}

\subsection{Counter-Intuitive Results of Bi-MTDP-A} Intuitively, implementing binarization on FP network inevitably induces representations degradation, as the gradient of the $sign$ function cannot be perfectly estimated~\cite{bengio2013estimating}. Thus, binarized models are impossible to outperform their full-precision counterpart models. However, \bimtdpA, a variant of \bimtdp~(\ie, full-precision backbone with only binarized multi-modal distillation) outperforms its fully FP version. Specifically, just binarizing the multi-modal distillation can simultaneously accelerate the model by $\sim$39\% and improve the mIoU for segmentation by $\sim$4\%, as shown in Table~\ref{table:nyudv2}. This result demonstrates that our method is not a naive trade-off between model performance and efficiency but a powerful tool for boosting multitask dense predictors. This exciting `free-lunch' achievement is even a bit of counter-intuitive. 
We speculate the reasons are that i) binarization on MMD can filter out task-irrelevant information; ii) and thus the information flow in the network is more effective. 
To testify this speculation, we conduct a series of experiments in two aspects, the representation ability of \bimtdpA~and information flow supplementation within the network. 

\noindent\textbf{Qualitative Study of Learned Features with \bimtdp}
To investigate the representation ability of \bimtdpA~and its FP counterparts, we visualize i) the feature maps behind the Binarized Multi-Modal Distillation (MMD) module in 2-D space via the t-SNE~\cite{van2008visualizing} algorithm, and ii) the regions where the network considers important via the Grad-Cam algorithm~\cite{selvaraju2017grad}. The results are shown in Fig.~\ref{fig:representation}. It is clear that binarized model, \bimtdpA~is able to filter the irrelevant information out via the binarized attention module (see Fig.~\ref{fig:representation} (a)), and thus helps learn more discriminative features (see Fig.~\ref{fig:representation} (b)) resulting in higher quantitative results. Overall, the generated spatial feature maps for segmentation are better. The enhanced representative ability can contribute to higher quantitative results.

\noindent\textbf{Analysis of Information Flow Supplementation within Network via Centered Kernel Alignment.} Analyzing distributional information flow within layers of neural networks is challenging because outputs of layers are distributed across a large number of neurons. Centered kernel alignment (CKA)~\cite{cortes2012algorithms,kornblith2019similarity,raghu2021vision} can address these challenges, by quantitatively comparing activations within or across networks. Specifically, for a network feed by $m$ samples, CKA algorithm takes $\mathbf{X} \in \mathbb{R}^{m\times p_1}$ and $\mathbf{Y} \in \mathbb{R}^{m\times p_2}$ as inputs which are output activations of two layers (with $p_1$ and $p_2$ neurons respectively). Letting $\mathbf{K} \triangleq \mathbf{X}\mathbf{X}^{\top}$
and $\mathbf{L} \triangleq \mathbf{Y}\mathbf{Y}^{\top}$ denote the Gram matrices for the two layers CKA computes:
\begin{equation}
    \mathrm{CKA}(\mathbf{K},\mathbf{L}) = \frac{\mathrm{HSIC}(\mathbf{K},\mathbf{L})}{\sqrt{\mathrm{HSIC}(\mathbf{K},\mathbf{K})\mathrm{HSIC}(\mathbf{L},\mathbf{L})}},
\end{equation}
where HSIC is the Hilbert-Schmidt independence criterion~\cite{gretton2007kernel}. Given the centering matrix $\mathbf{H}=\mathbf{I}_n - \frac{1}{n}\mathbf{1}\mathbf{1}^{\top}$ and the centered Gram matrices $\mathbf{K}^{\prime}=\mathbf{H}\mathbf{K}\mathbf{H}$ and $\mathbf{L}^{\prime}=\mathbf{H}\mathbf{L}\mathbf{H}$, $\mathrm{HSIC}=\frac{\mathrm{vec}(\mathbf{K}^{\prime})\mathrm{vec}(\mathbf{L}^{\prime})}{(m-1)^2}$, the similarity between these centered Gram matrices. Importantly, CKA is invariant to orthogonal transformation of representations (including permutation of neurons), and the normalization term ensures invariance to isotropic scaling. These properties enable meaningful comparison and analysis of neural network hidden representations. 

\begin{table*}[!h]\footnotesize
\centering
\caption{Results on NYUD-v2. \bimtdpA: only implementing the binarization in the multi-modal distillation module, \bimtdpB: fully-binarized model.} 
\scalebox{1.0}{
\begin{tabular}{cc|cccc|ccc}\toprule
     \multicolumn{2}{c|}{Model}  & \multirow{2}{*}{SemSeg $\uparrow$} & \multirow{2}{*}{Depth $\downarrow$} & \multirow{2}{*}{Normal $\downarrow$} & \multirow{2}{*}{Bound $\uparrow$} & float32 & binary & FLOPs \\ 
   Method & Backbone & & & & & Params (M) & Params (M) & (G)\\ \midrule
              Cross~\cite{misra2016cross}  &   & 36.34        & 0.629              & 20.88  & 76.38 & 241.46  & 0 & 338.09\\ 
              PAP~\cite{zhang2019pattern}  &   & 36.72             & 0.617              & 20.82  & 76.42 & 189.10 & 0 & 256.86\\ 
              PSD~\cite{zhou2020pattern}  &   & 36.69             & 0.625              & 20.87  & 76.42 & 224.67  & 0 & 315.60 \\ 
              PAD~\cite{xu2018pad}   & CNN & 36.61             & 0.627              & 20.85  & 76.38 & 170.98 & 0 & 230.91 \\ 
              MTI~\cite{vandenhende2020mti}  &   & 45.70             & 0.537              & 20.27  & 77.86 & 144.87 & 0 &212.98\\ 
              ATRC~\cite{bruggemann2021exploring}  &  & 45.87             & 0.540              & 20.09  & 77.34 & 180.00 & 0 &249.24\\ 
              \rowcolor{brown!20} MTI + \bimtdpB &  &  39.20   & 0.612     & 21.04  & 76.86 & 6.45 & 138.42 & 18.07\\
              \rowcolor{brown!20} MTI + \bimtdpA &  &  47.71    &  0.530     &  20.06  & 77.36 & 90.28 & 54.59 & 130.65\\\midrule
               InvPT~\cite{ye2022inverted} & ViT-B  &   50.30    &   0.536     &  19.00  & 77.60 & 154.79 & 0 & 244.71\\
               \rowcolor{brown!20} InvPT + \bimtdpA &  &   51.20\textcolor{green}{\scriptsize (0.90$\uparrow$)}    &  0.528\textcolor{green}{\scriptsize (0.08$\downarrow$)}     &  19.50\textcolor{red}{\scriptsize (0.50$\uparrow$)}  & 77.68\textcolor{green}{\scriptsize (0.08$\uparrow$)} & 127.92 & 26.87 & 183.81\\\cline{3-9}
               InvPT~\cite{ye2022inverted} & ViT-L &  53.56    &  0.518     &  19.04  & 78.10 & 239.22 & 0 & 331.60\\
               \rowcolor{brown!20}{3-9} InvPT + \bimtdpA &  &  54.86\textcolor{green}{\scriptsize (1.30$\uparrow$)}    &  0.515\textcolor{green}{\scriptsize (0.03$\downarrow$)}     &  19.50\textcolor{red}{\scriptsize (0.46$\uparrow$)}  & 78.20\textcolor{green}{\scriptsize (0.10$\uparrow$)} & 212.34 & 26.87 & 301.88\\
              \bottomrule
\end{tabular}}
\vspace{-0.25in}
\label{table:nyudv2}
\end{table*}

Therefore, we introduce CKA to study the information flow in the multitask dense prediction models. 
In the heat-map, the lighter the dot, the more similar the two corresponding layers. Higher similar score between two layers' output representations means those two layers share more information. The results are presented in Fig.~\ref{fig:representation} (c), we can see that the similar scores among front layers and back layers in \bimtdpA~is much higher than the ones in MTI-Net~\cite{vandenhende2020mti}. 
This indicates that \bimtdpA~is able to supplement information flow within network, and thus boost the model performance.

\section{Experiments}

In this section, we conducted comprehensive experiments to evaluate our proposed method on two datasets for dense prediction tasks: PASCAL Context~\cite{everingham2010pascal} and NYUD-v2~\cite{silberman2012indoor}. 
We first describe the implementation details of \bimtdp, and then compare our method with SoTA binary neural networks in the task of object detection to demonstrate superiority of the proposed method. Finally, we validate the effectiveness of information bottleneck and feature-based knowledge distillation by a series of ablative studies. 

\subsection{Datasets, Evaluation, Implementation Details}
\noindent \textbf{Datasets.} \textbf{\textit{PASCAL-Content}} is a popular dataset for dense prediction tasks. We use the split from PASCAL-Context which has annotations for semantic segmentation, human part segmentation, semantic edge detection, surface normals prediction and saliency detection. Note that some annotations were distilled by \cite{maninis2019attentive} using pre-trained SoTA models~\cite{chen2018encoder}. \textbf{\textit{NYUD-v2}} contains various indoor scenes such as offices and living rooms with 795 training and 654 testing images. It provides different dense labels, including semantic segmentation, monocular depth estimation, surface normal estimation and object boundary detection. 

\begin{table*}[!t]\footnotesize
\centering
\caption{Results on PASCAL-VOC. \bimtdpA: only implementing the binarization in the multi-modal distillation module, \bimtdpB: fully-binarized model.} 
\scalebox{0.95}{
\begin{tabular}{cc|ccccc|ccc}\toprule
          \multicolumn{2}{c|}{Model}   & \multirow{2}{*}{SemSeg $\uparrow$} & \multirow{2}{*}{Parsing $\uparrow$} & \multirow{2}{*}{Saliency $\uparrow$} & \multirow{2}{*}{Normal $\downarrow$} & \multirow{2}{*}{Bound $\uparrow$} & float32 & binary & FLOPs \\ 
   Method & Backbone  & & & & & & Params (M) & Params (M) & (G)\\\hline
              ASTMT~\cite{maninis2019attentive}  & & 68.00        & 61.10              & 65.70  & 14.70 & 72.40 & 364.72 & 0 & 501.27\\ 
              PAD~\cite{xu2018pad}  &   & 53.60             & 59.60              & 65.80  & 15.30 & 72.50 & 231.80 & 0 & 289.46\\ 
              MTI~\cite{vandenhende2020mti}  &   & 61.70             & 60.18              & 84.78  & 14.23 & 70.80 & 218.56 & 0 & 280.12\\ 
              ATRC~\cite{bruggemann2021exploring}  &  \multirow{2}{*}{CNN} & 62.69             & 59.42              & 84.70  & 14.20 & 70.96 & 241.45 & 0 & 310.19\\ 
              ATRC-A~\cite{bruggemann2021exploring}  &   & 63.60             & 62.23              & 83.91  & 14.30 & 70.86  & 249.87 & 0 & 320.57\\ 
              ATRC-B\cite{bruggemann2021exploring} &   & 67.67             & 62.93              & 82.29  & 14.24 & 72.42 & 280.01 & 0 & 383.21\\ 
              \rowcolor{brown!20} MTI~\cite{vandenhende2020mti} + \bimtdpB   &  &   48.10    & 56.28   & 64.42  & 14.29   & 76.89 & 13.67 & 204.89 & 31.38\\     
              \rowcolor{brown!20} MTI~\cite{vandenhende2020mti} + \bimtdpA   &  &   62.98\textcolor{green}{\scriptsize (1.28$\uparrow$)}     & 60.44\textcolor{green}{\scriptsize (0.26$\uparrow$)}     & 83.56\textcolor{red}{\scriptsize (1.22$\downarrow$)}  & 14.31\textcolor{red}{\scriptsize (0.08$\uparrow$)}   & 71.28\textcolor{green}{\scriptsize (0.26$\uparrow$)} & 153.71 & 64.85 & 194.51\\\hline              
             InvPT\cite{ye2022inverted}   &   ViT-B  & 77.50             & 66.83              & 83.65  &  14.63 & 73.00 & 176.35  & 0 & 274.68\\ 
             \rowcolor{brown!20} InvPT\cite{ye2022inverted}  + \bimtdpA  &    &    76.84 \textcolor{red}{\scriptsize (0.66$\downarrow$)}            & 67.10\textcolor{green}{\scriptsize (0.27$\uparrow$)}              & 84.97 \textcolor{green}{\scriptsize (1.32$\uparrow$)}  & 13.69\textcolor{green}{\scriptsize (0.94$\downarrow$)} & 73.04 \textcolor{green}{\scriptsize (0.04$\uparrow$)}  & 154.68 & 21.67 & 220.76 \\ \cline{3-10}
             InvPT\cite{ye2022inverted}   &   ViT-L  & 79.03             & 67.61              & 84.81  & 14.15 & 73.00 & 422.93  & 0 & 425.37\\ 
             \rowcolor{brown!20} InvPT\cite{ye2022inverted} + \bimtdpA  &    &   79.83 \textcolor{green}{\scriptsize (0.80$\uparrow$)}            & 68.17\textcolor{green}{\scriptsize (0.56$\uparrow$)}              & 84.92 \textcolor{green}{\scriptsize (0.11$\uparrow$)}  & 13.92\textcolor{green}{\scriptsize (0.23$\downarrow$)} & 73.03 \textcolor{green}{\scriptsize (0.03$\uparrow$)} & 401.26  & 21.67 & 382.68\\\bottomrule
\end{tabular}}
\vspace{-0.2in}
\label{table:pascal}
\end{table*}

\noindent \textbf{Evaluation.} Semantic segmentation (Semseg) and human parsing (Parsing) are evaluated with mean Intersection over Union (mIoU); monocular depth estimation (Depth) is evaluated with Root Mean Square Error (RMSE); surface normal estimation (Normal) is evaluated by the mean error (mErr) of predicted angles; saliency detection (Saliency) is evaluated with maximal F-measure (maxF); object boundary detection (Boundary) is evaluated with the optimal-dataset-scale F-measure (odsF). To evaluate the model efficiency \textit{w.r.t.} memory cost and inference speed, we adopt the number of parameters and FLOPs for a single round of the model inferring an input image.

\noindent \textbf{Implementation details.} We build our approach on the most prevalent backbone architecture, \textit{i.e.}, HRNet as previous SoTA methods~\cite{vandenhende2020mti,bruggemann2021exploring}. The task-specific heads are also implemented as two basic residual blocks, \textit{i.e.}, binarized BasicBlock and binarized Bottleneck with additional binary shortcuts as BiReal-Net~\cite{liu2020bi}. We use $\ell_1$ loss for depth estimation and cross-entropy loss for semantic segmentation on NYUD-v2. As in the prior work, the edge detection task is trained with a positive weighted wpos = 0:95 binary cross-entropy loss. We do not adopt a particular loss weighing strategy on NYUD-v2, but simply sum the losses together. On PASCAL, we reuse the training setup from~\cite{vandenhende2020mti} to facilitate a fair comparison. We reuse the loss weights from there. The initial task predictions in the front-end of the network use the same loss weighing as the final task predictions. We refer to the supplementary material for further implementation details. Importantly, we use Adam optimizer~\cite{kingma2014adam} in training, but with different learning rates for binary parameters (1e-5) and FP parameters (1e-4), as Adam with a larger learning rate for binary parameters can lead to better training results~\cite{liu2021adam}. Note that our project is based on the codebases for MTI-Net~\cite{vandenhende2020mti} and more details can be found in the \textbf{codes} in the \textbf{Supplemental Materials}.

\subsection{Comparison with State-of-the-Art}

Tabs. \ref{table:nyudv2} and \ref{table:pascal} present a comparative analysis of the proposed Binary Multitask Dense Predictor (\bimtdp) with current state-of-the-art models on the NYUD-v2 and PASCAL-Context datasets. This comparison includes notable CNN-based methods such as PAD-Net~\cite{xu2018pad}, ASTMT~\cite{maninis2019attentive}, MTI-Net~\cite{vandenhende2020mti}, and ATRC~\cite{bruggemann2021exploring}, among others. \bimtdp~demonstrates exceptional performance, outperforming other models in 6 of the 9 evaluated metrics, particularly in complex scene understanding tasks like Semantic Segmentation and Parsing. Remarkably, on the NYUD-v2 benchmark, \bimtdp~surpasses the previously best-performing CNN-based method (ATRC) by a margin of +1.8 (mIoU) in Semantic Segmentation, while requiring only 62\% of the storage space for weights and 56\% of the computational FLOPs.

Furthermore, the application of \bimtdp~to the ViT-based state-of-the-art method, InvPT~\cite{ye2022inverted}, showcases \bimtdp's ability to enhance model performance while also improving efficiency. This demonstrates the broad applicability and generalization capability of \bimtdp~across different architectural frameworks.

\begin{figure}[htbp]
\centering
  \includegraphics[width=0.48\textwidth]{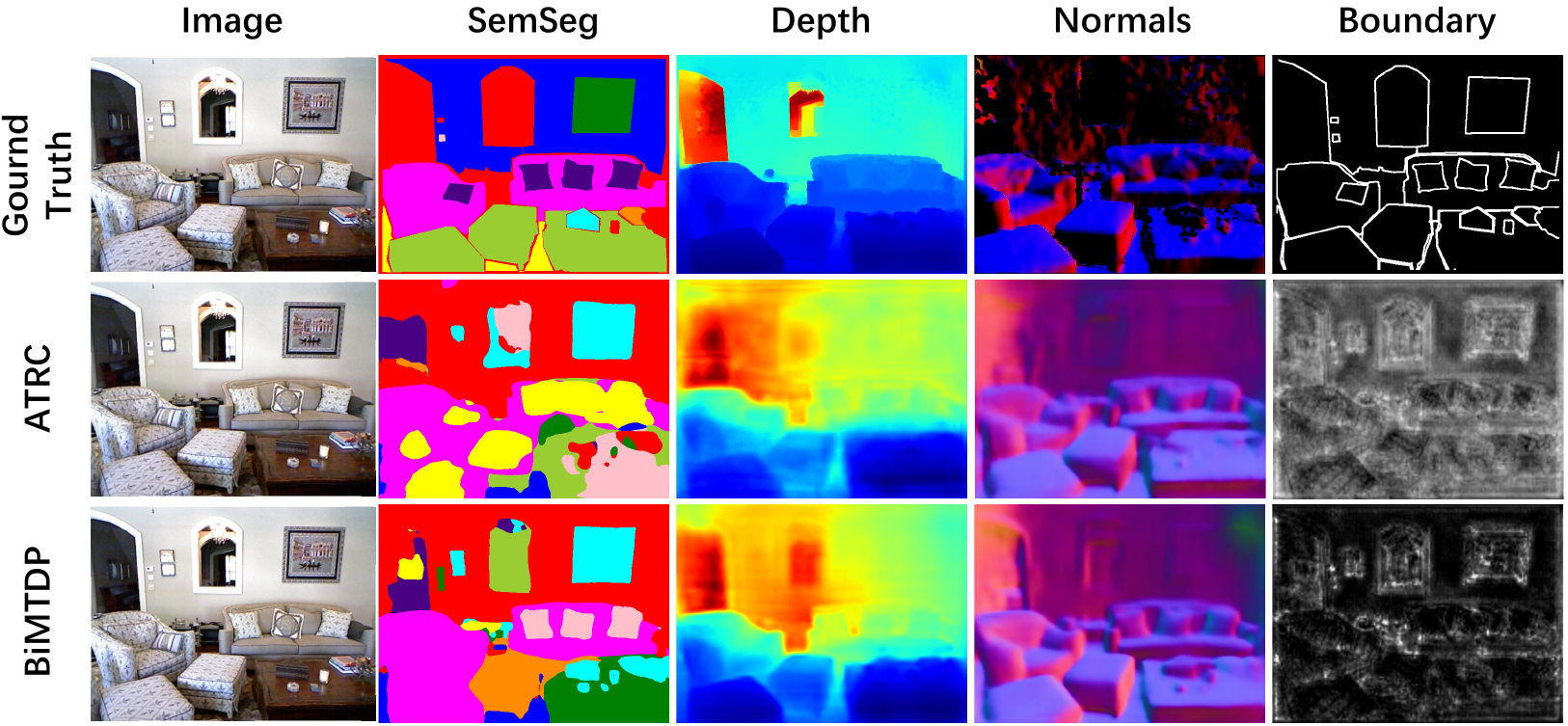}
  \captionsetup{font=small}
  \caption{Qualitative comparison with ATRC on NYUD-v2.}
  \vspace{-0.1in}
  \label{fig:visualization_pascal}
\end{figure}

\begin{figure}[htbp]
\centering
  \includegraphics[width=0.48\textwidth]{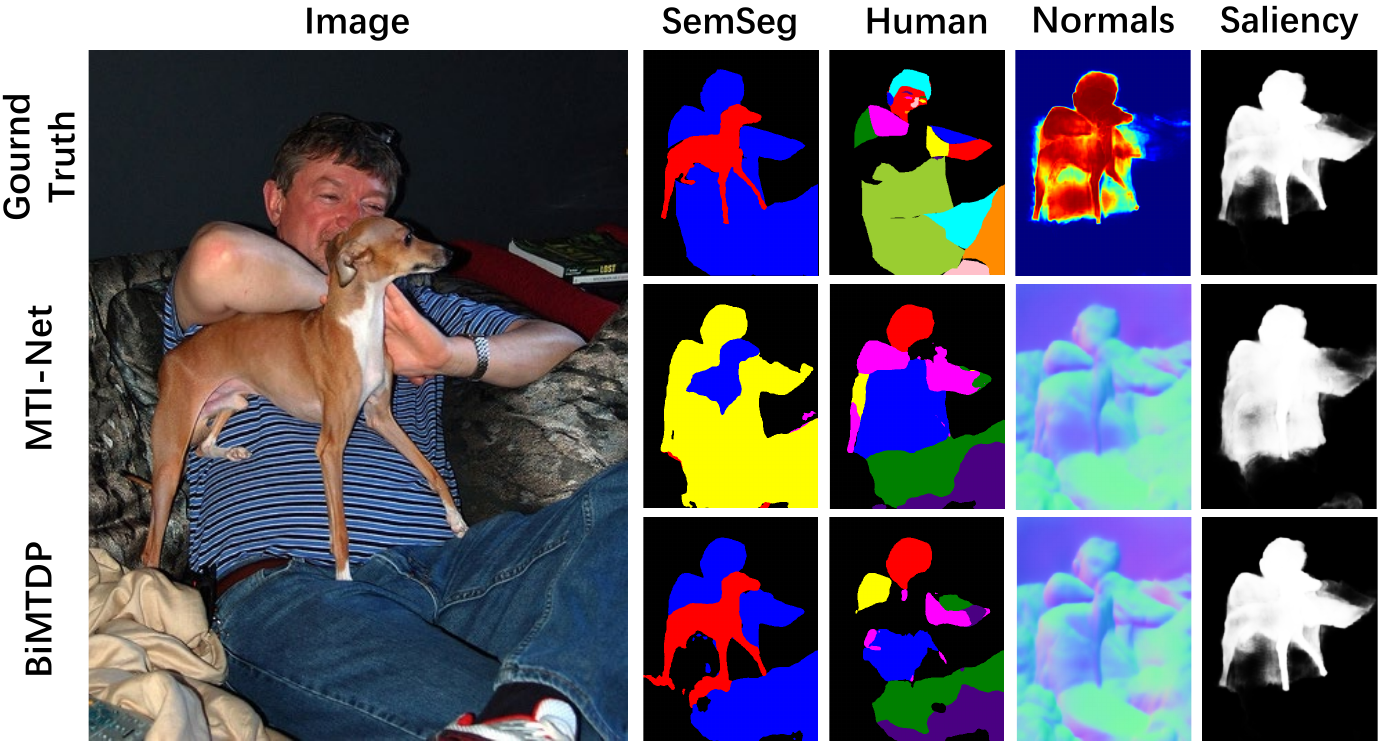}
  \captionsetup{font=small}
  \caption{Qualitative comparison with MTI-Net on PASCAL-Context.}
  \vspace{-0.2in}
  \label{fig:visualization_nyud}
\end{figure}

For a qualitative assessment, Figs.~\ref{fig:visualization_nyud} and \ref{fig:visualization_pascal} displays prediction examples from various models. These examples illustrate that \bimtdpA~not only competes with but occasionally surpasses the state-of-the-art ATRC in qualitative performance. 

\subsection{Ablative Studies}
Tab.~\ref{table:ablation} shows a series of ablative studies of \bimtdpA~and \bimtdpB~with an HRNet-48 backbone on NYUD-v2. We verify how different components of our model contribute to the multi-task improvements. In summary, every designed module positively impacts the overall model. We would like to highlight the main intuition of our method that binarized dense prediction models in a multitask manner largely outperform the binarized single models through \bimtdpB~\emph{w.o.} VIB \& KD \vs Bi-Single. 
\vspace{0.1in}


\hspace{-0.5cm}
\begin{minipage}{0.49\textwidth}
\centering
\captionof{table}{Ablative studies on NYUD-v2.} 
\scalebox{0.7}{
\begin{tabular}{l|cccc}\toprule
     \multirow{2}{*}{Model}   & \multirow{2}{*}{SemSeg $\uparrow$} & \multirow{2}{*}{Depth $\downarrow$} & \multirow{2}{*}{Normal $\downarrow$} & \multirow{2}{*}{Bound $\uparrow$}  \\ 
     & & & & \\\hline

            \rowcolor{brown!20} \bimtdpA & \textbf{47.71}    & \textbf{0.530}     & \textbf{20.06}  & \textbf{77.36} \\
              \bimtdpA~w.o. binary MMD & 45.70             & 0.537              & 20.27  & 77.86\\\hline
            \rowcolor{brown!20} \bimtdpB & \textbf{39.20}   & \textbf{0.612}     & \textbf{21.04}  & \textbf{76.86} \\
              \bimtdpB~w.o. VIB & 35.04   & 0.652     & 22.31  & 73.88 \\
              \bimtdpB~w.o. KD & 36.20   & 0.640     & 21.99  & 74.86 \\
              \bimtdpB~w.o. VIB \& KD & 33.99   & 0.667     & 23.24  & 71.30 \\
              Bi-Single   & 16.20        & 0.872              & 29.36  & 69.24 \\ 
             \bottomrule
\end{tabular}}
\label{table:ablation}
\end{minipage}



\section{Conclusion}
\label{sec:con}
In this paper, we significantly accelerate cumbersome dense prediction models, in which  BNNs for relevant tasks are modeled and optimized under a multitask framework to supplement  degraded information caused by binarization operations. Based on this binary baseline, we further introduce variational information bottleneck and feature-based knowledge distillation to supplement information flow. Experiment results show that our method significantly accelerates existing SoTA methods with comparably small performance drop over the mainstream dense prediction tasks on PASCAL VOC and NYUD-v2. Intriguingly, \bimtdp~not only reaches SoTA \wrt performance but also saves computational costs, compared with SoTA method ARTC~\cite{bruggemann2021exploring}. 

\noindent \textbf{Acknowledgments:} This research is partially supported by NSF IIS-2309073, ECCS-212352101 and Cisco unrestricted gift. This article solely reflects the opinions and conclusions of its authors and not the funding agencies.

\clearpage


{\small
\bibliographystyle{ieee_fullname}
\bibliography{egbib}
}


\end{document}


\title{Efficient Multitask Dense Predictor via Binarization}

\author{First Author\\
Institution1\\
Institution1 address\\
{\tt\small firstauthor@i1.org}
\and
Second Author\\
Institution2\\
First line of institution2 address\\
{\tt\small secondauthor@i2.org}
}
\maketitle

\section{Appendix}

\subsection{Derivative of VIB}
The information bottleneck (IB) theory closely relates to compression with the hypothesis that the data misfit and the model complexity should simultaneously be minimized, so that the redundant information irrelevant to the task is exclusive in the compressed model and the capacity of the lightweight model is fully utilized. The dense prediction tasks can be regarded as a Markov process with the following Markov chain:
\begin{equation}
    X\xrightarrow{}F\xrightarrow{}O,
\end{equation}
where $X$ means the input images, $F$ stands for the high-level feature maps output by the backbone part, and $O$ represent the final task-specific output of the binary model. According to the Markov chain, the objective of the IB principle is written as follows:
\begin{equation}
    \min_{\theta}I(X; F) -\beta\cdot I(F;O),
\end{equation}
where $\theta$ is the parameters of the binary model, and $I(X; F)$ represents the mutual information between two random variables $X$ and $F$. The mutual information is formalized as follows:
\begin{equation}
    I(X; F) = \mathbb{E}_{\mathbf{x}\sim p(\mathbf{x})}\mathbb{E}_{\mathbf{f}\sim p(\mathbf{f}\mid \mathbf{x})}\log\frac{p(\mathbf{f}\mid \mathbf{x})}{p(\mathbf{f})},
\end{equation}
where $\mathbf{x}$ and $\mathbf{f}$ are the specific input images and the corresponding high-level feature maps. $p(\mathbf{x})$ and $p(\mathbf{f})$ are the prior distribution of $\mathbf{x}$ and $\mathbf{x}$. $p(\mathbf{f}\mid \mathbf{x})$ is the posterior distribution of the high-level feature map conditioned on the input. 
Minimizing the mutual information between the images and the high-level feature maps constrains the amount of information that the detector extracts, and maximizing the mutual information between the high-level feature maps and the following tasks enforces the task-specific heads to preserve more information related to the task. As a result, the redundant information irrelevant \wrt final tasks is removed. 

We parameterize $p(\mathbf{f}\mid \mathbf{x})$ by the backbone due to its intractability, where evidence-lower-bound (ELBO) minimization is applied for relaxation. To estimate $I(X; F)$, we sample the training set to obtain the image $\mathbf{x}$ and sample the distribution parameterized by the backbone to acquire the corresponding high-level feature map $\mathbf{f}$. 

Similar to above equation, we rewrite the mutual information between the high-level feature maps and the final output as follows:
\begin{equation}
    I(F; O) = \mathbb{E}_{\mathbf{f}\sim p(\mathbf{f}\mid \mathbf{x})}\mathbb{E}_{\mathbf{o}\sim p(\mathbf{o}\mid \mathbf{f})}\log\frac{p(\mathbf{o}\mid \mathbf{f})}{p(\mathbf{o})}.
\end{equation}
Combining above equations and Deep variational information bottleneck (ICLR, 2017), we can derive the VIB-based objective function can be formulated as a term of information loss, written as follows:
\begin{equation}
    \mathcal{L}_{vib} = \textit{KL}\left[p(\mathbf{Z}\mid \mathbf{F_{B}}), r(\mathbf{Z}) \right],
\end{equation}
where $\mathbf{F_{B}}$ is the input binary backbone representation,  $\mathbf{Z}$ is the latent representation variable, $p(\mathbf{Z}\mid \mathbf{A_{B}})$ is a multivariate Gaussian distribution, and $r(\mathbf{Z})$ is a standard normal distribution. Generally, the latter is a regularization term controlling how much information of the input is filtered out.


\subsection{Implementation Details and Code}
\noindent\textbf{How to run the codes.}

\noindent\textbf{Step 1:} Download a pre-trained hrnet48 backbone and store it  in /Bi-MTDP/models/pretrained\_models

\noindent\textbf{Step 2:} Run the training and testing file on  /Bi-MTDP/main.py

{\small
\bibliographystyle{ieee_fullname}
\bibliography{egbib}
}